\documentclass[runningheads]{llncs}

\usepackage[T1]{fontenc}
\usepackage{graphicx,verbatim}
\usepackage[inline]{enumitem}
\usepackage{booktabs}
\usepackage{multirow}
\usepackage{amssymb} 
\usepackage{amsmath} 
\usepackage{siunitx}
\usepackage{marvosym}

\usepackage{color}
\usepackage[colorlinks=true,
    linkcolor=blue,
    citecolor=blue,
    urlcolor=blue]{hyperref}

\begin{document}
\title{Beyond Visual Cues: CoT-Enhanced Reasoning for Semi-supervised Medical Image Segmentation}
%

\author{
Yuming Chen\inst{1,2}
\and Yuxin Xie\inst{1,2}
\and Tao Zhou\inst{3}
\and Yi Zhou\inst{1,2}\textsuperscript{(\Letter)}
}
\authorrunning{Chen et al.}
\institute{
School of Computer Science and Engineering,
Southeast University, Nanjing, China
\and
Key Laboratory of New Generation Artificial Intelligence Technology and
Its Interdisciplinary Applications,
Ministry of Education, Nanjing, China
\and
Nanjing University of Science and Technology,
Nanjing, China
\\
\email{\{220252314@seu.edu.cn, yizhou.szcn@gmail.com\}}
}
  
\maketitle              
\begin{abstract}
Semi-supervised medical image segmentation has emerged as a dominant research problem in medical image analysis, mitigating annotation scarcity by leveraging consistency regularization on unlabeled data. However, existing approaches operate predominantly via visual pattern matching, relying heavily on pixel-level similarities. This visual-centric dependency often falters in clinical scenarios characterized by the visual-semantic mismatch, where visually similar lesions warrant distinct diagnostic conclusions, thus failing to capture the underlying diagnostic logic used by experts. To address this, we move beyond visual cues and propose CERS (CoT-Enhanced Reasoning Segmentation), a framework that integrates Chain-of-Thought (CoT) reasoning to distinguish pathologically distinct cases. Specifically, we construct a knowledge pool enriched with linguistic reasoning descriptions generated by large language models (LLMs). A semantic-aware reference selection strategy is introduced to identify historical evidence, filtering candidates first by morphology, and then refining them via CoT consistency to eliminate hard negatives. Furthermore, a multi-scale coordinate attention module (MCAM) is designed to effectively fuse this reasoning-derived context into the decoding process. Extensive experiments demonstrate the superiority of CERS against state-of-the-art approaches, particularly in resolving boundary ambiguities and semantic inconsistencies. The code is available at \url{https://github.com/cymasuna/CERS}.


\end{abstract}
\section{Introduction}
Medical image segmentation is a cornerstone of computer-aided diagnosis, enabling precise delineation of organs and lesions~\cite{khan2025comprehensive,wang2022medical}. However, the efficacy of deep learning models heavily relies on large-scale, high-quality annotations, which are difficult and laborious to obtain~\cite{han2024deep}. To address this, Semi-Supervised Learning (SSL) has emerged as a widely adopted paradigm, leveraging limited labeled data alongside abundant unlabeled data~\cite{han2024deep,song2023comprehensive}. Although existing SSL approaches, ranging from consistency regularization~\cite{tarvainen2017mean,bai2023bidirectional,wu2025dual,li2023lvit} to pseudo label supervision~\cite{shen2023co,zeng2024consistency}, have achieved promising results, they predominantly focus on pixel-level visual features.

A fundamental limitation of current SSL methods is their reliance on visual pattern matching. Although effective for standard cases, this approach often fails when faced with ambiguous boundaries, low-contrast lesions, imaging artifacts caused by respiration, or anatomically similar structures that mimic pathological appearance, leading to false positives or missed detections~\cite{zeng2025exploring}. In clinical practice, physicians overcome such challenges by recalling similar historical cases and applying diagnostic reasoning to infer lesion boundaries beyond purely visual evidence~\cite{welter2011towards}. Following this observation, recent studies have suggested that introducing external reference cues can significantly enhance model performance~\cite{zhao2025retrieval}. However, a critical challenge remains that clinical scenarios often exhibit a visual-semantic mismatch, where images with high visual similarity warrant distinct radiological reading conclusions~\cite{fang2021combating}. Consequently, existing text-driven segmentation methods, which rely solely on superficial and incomplete medical text descriptions, often fail to capture the actual radiological reading logic~\cite{li2023lvit,xie2024simtxtseg,xie2026core}. This highlights the necessity to incorporate high-level reasoning capabilities to distinguish pathologically distinct cases that appear visually similar.

To address this challenge, we propose CERS (CoT-Enhanced Reasoning Segmentation), a framework that integrates Chain-of-Thought (CoT)~\cite{wei2022chain} reasoning into the semi-supervised segmentation task. Instead of relying on unreliable visual matching, CERS generates linguistic reasoning descriptions to capture the semantic context of lesions. Using these CoT descriptions, our method performs a semantic-aware retrieval that bridges the gap between visual appearance and diagnostic logic. This ensures that the utilized historical references are not just visually similar, but logically consistent with the target case. These reasoning-guided cues are then effectively injected into the segmentation network to resolve ambiguities in unlabeled data.

\textbf{The main contributions are summarized as follows.}
\begin{enumerate*}[label=\textbf{(\arabic*)}, itemjoin=\quad]
    \item We propose CERS, a CoT-enhanced framework that generates segmentation-aware reasoning descriptions to provide high-level semantic guidance for semi-supervised segmentation.
    \item We design a Multi-scale Coordinate Attention Module (MCAM) with a dual-decoder consistency scheme to effectively fuse reasoning-derived context into the decoder, significantly improving segmentation accuracy in challenging scenarios.
    \item We conduct extensive experiments on multiple public datasets, showing that CoT-enhanced reasoning moves segmentation beyond purely visual cues and significantly improves semi-supervised performance.
\end{enumerate*}

\section{Method}
\begin{figure}[t]
  \centering
    \includegraphics[width=\linewidth]{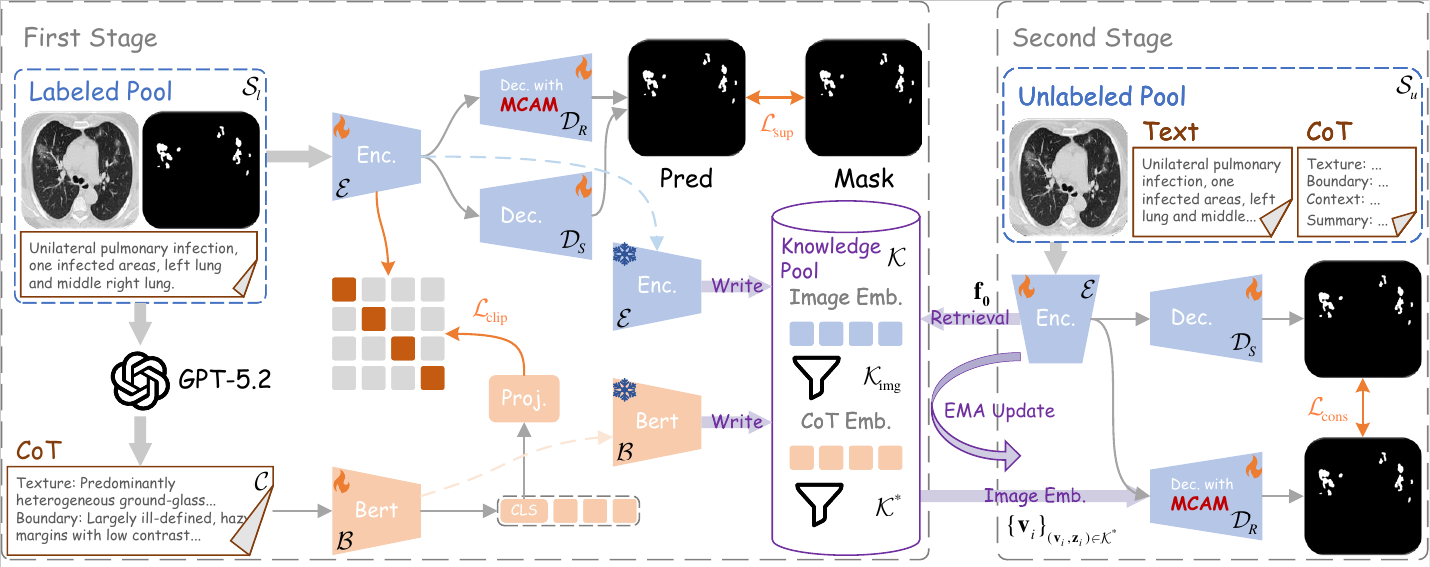}
    \caption{Overview of CERS. 
    A frozen LLM generates CoTs, and labeled CoTs form a knowledge pool for image–CoT hybrid retrieval. Retrieved context is fused with image features via MCAM and used by a retrieval-aware decoder to enable reasoning-guided semi-supervised segmentation beyond visual cues.
    }
    \label{fig:fig1a}
\end{figure}

We study semi-supervised medical image segmentation with a labeled set $\mathcal{S}_l=\{(x_i^l,y_i^l,t_i^l)\}_{i=1}^{N_l}$, where $x_i^l\in\mathbb{R}^{H\times W\times C}$ is a medical image, $y_i^l\in\{0,1\}^{H\times W}$ is the corresponding mask and $t_i^l$ is the associated text, and an unlabeled set $\mathcal{S}_u=\{(x_j^u,t_j^u)\}_{j=1}^{N_u}$. For every sample, we generate a Chain-of-Thought (CoT) via a frozen LLM, denoted $c_i^l$ for labeled and $c_j^u$ for unlabeled samples. \textbf{Only labeled CoTs} are used to populate the knowledge pool $\mathcal{K}$, avoiding leakage from unlabeled data. As illustrated in Fig.~\ref{fig:fig1a}, our multi-modal CERS model is $f_\theta=\{\mathcal{E},\mathcal{D}_S,\mathcal{D}_R\}$, where $\mathcal{E}$ is a shared encoder, $\mathcal{D}_S$ a standard segmentation decoder and $\mathcal{D}_R$ a retrieval-aware decoder that fuses retrieved context. CERS employs an image–CoT hybrid retrieval in two stages: a preliminary screening retrieves visually similar candidates to ensure morphological consistency, and a fine-grained re-ranking uses CoT similarity to remove hard negatives that are visually similar but semantically distinct. Retrieved features are fused with the the current features via the Multi-scale Coordinate Attention Module to yield a retrieval-aware enhanced representation, which is consumed by $\mathcal{D}_R$.
\subsection{Segmentation-Aware Reasoning Knowledge Pool Construction}
\subsubsection{CoT Generation.}
We generate Chain-of-Thought (CoT) descriptions for each sample by prompting GPT-5.2~\cite{singh2025openai}. 
For labeled samples we provide the image $x_i^l$, its mask $y_i^l$ (as a translucent overlay), and text $t_i^l$; for unlabeled samples we provide only the image $x_j^u$ and text $t_j^u$.
The resulting CoTs for labeled and unlabeled sets are denoted by $\mathcal{C}^l$ and $\mathcal{C}^u$ respectively.




Prompts elicit four diagnostic aspects—\emph{texture}, \emph{boundary}, \emph{context}, and a concise \emph{summary}—thereby encouraging the LLM to produce semantically discriminative cues for segmentation. 
We omit explicit spatial coordinates because free-text locations are often ambiguous; morphology-aware image embeddings already supply spatial priors.
To verify the reliability of the generated prompts, we randomly sampled 100 cases for expert manual review, which confirmed the high clinical accuracy (more than 95\%) of the LLM-generated CoT.

\subsubsection{Warm-up  to Build Knowledge Pool.}
To initialize the model and ensure stable retrieval in the subsequent SSL phase, we first conduct a supervised warm-up (Eq.~\ref{eq2}) using \textbf{only the labeled set $\mathcal{S}_l$}. During this phase the multi-modal model $f_\theta=\{\mathcal{E},\mathcal{D}_S,\mathcal{D}_R\}$ is trained on labeled samples to learn reliable segmentation and cross-modal embeddings. The supervised segmentation loss is defined as:
\begin{equation}
\mathcal{L}_{\mathrm{sup}}=\frac{1}{|\mathcal{S}_l|}\sum_{(x,y,t)\in\mathcal{S}_l}\Big[\mathcal{L}_{\mathrm{seg}}\big(\mathcal{D}_S(\mathcal{E}(x, t)),y\big)+\mathcal{L}_{\mathrm{seg}}\big(\mathcal{D}_R(\mathcal{E}(x, t)),y\big)\Big],
\end{equation}
where $\mathcal{L}_{\mathrm{seg}}$ combines standard cross-entropy loss and Dice loss. 
We obtain an image embedding $\mathbf{v}_i=\mathcal{E}(x_i^l, t_i^l)$
and a CoT embedding $\mathbf{z}_i=\mathcal{B}(c_i^l)$ using a texture encoder $\mathcal{B}$ and apply contrastive objective~\cite{radford2021learning} $\mathcal{L}_{\mathrm{clip}}$ to align these embeddings. The overall warm-up loss is defined as:
\begin{equation}
\label{eq2}
\mathcal{L}_{\mathrm{warm}}=\mathcal{L}_{\mathrm{seg}}+\lambda_{\mathrm{clip}}\mathcal{L}_{\mathrm{clip}},
\end{equation}
where $\lambda_{\mathrm{clip}}$ balances segmentation supervision and cross-modal alignment. After warm-up convergence, the trained image encoder $\mathcal{E}$ and texture encoder $\mathcal{B}$ are fixed and used to encode all labeled samples. These paired representations are then stored in the knowledge pool
$
\mathcal{K}=\{(\mathbf{v}_i,\mathbf{z}_i)\}_{i=1}^{N_l},
$
which serves as the retrieval database in the subsequent semi-supervised learning stage.

\subsection{Image–CoT Joint Cues}

In the second training stage, we continue optimization based on the warm-up model and leverage the constructed knowledge pool $\mathcal{K}$ to retrieve reliable cues for both labeled and unlabeled samples. 
Given an input sample $x$ and the associated text with its generated CoT $c$, we first extract the image embedding $\mathbf{v}=\mathcal{E}(x,t)$ and the CoT embedding $\mathbf{z}=\mathcal{B}(c)$. CERS performs retrieval in a coarse-to-fine manner:
\begin{enumerate*}[label=\textbf{(\arabic*)}, itemjoin=\quad]
    \item \textbf{Image-based coarse filtering}: To ensure visual consistency, we conduct an image-based coarse retrieval over the knowledge pool using cosine similarity, 
    $s_i^{\mathrm{img}}=\cos(\mathbf{v},\mathbf{v}_i)$, where $(\mathbf{v}_i,\mathbf{z}_i)\in\mathcal{K}$. 
    The top-$K$ samples with the highest $s_i^{\mathrm{img}}$ are selected to form a candidate set $\mathcal{K}_{\mathrm{img}}$, which contains morphologically similar samples.
    \item \textbf{CoT-based fine re-ranking}: Among the candidates in $\mathcal{K}_{\mathrm{img}}$, we further compute CoT similarity to refine retrieval results, $s_i^{\mathrm{cot}}=\cos(\mathbf{z},\mathbf{z}_i)$, where $(\mathbf{v}_i,\mathbf{z}_i)\in\mathcal{K}_{\mathrm{img}}$. The candidates are re-ranked according to $s_i^{\mathrm{cot}}$, and the top-$M$ samples are retained as the final retrieved evidence set $\mathcal{K}^\ast$. This fine-grained re-ranking effectively filters out \emph{hard negatives} that are visually similar but diagnostically inconsistent with the query.
\end{enumerate*}


\subsection{Multi-scale Coordinate Attention Module}
\begin{figure}[t]
  \centering
    \includegraphics[width=\linewidth]{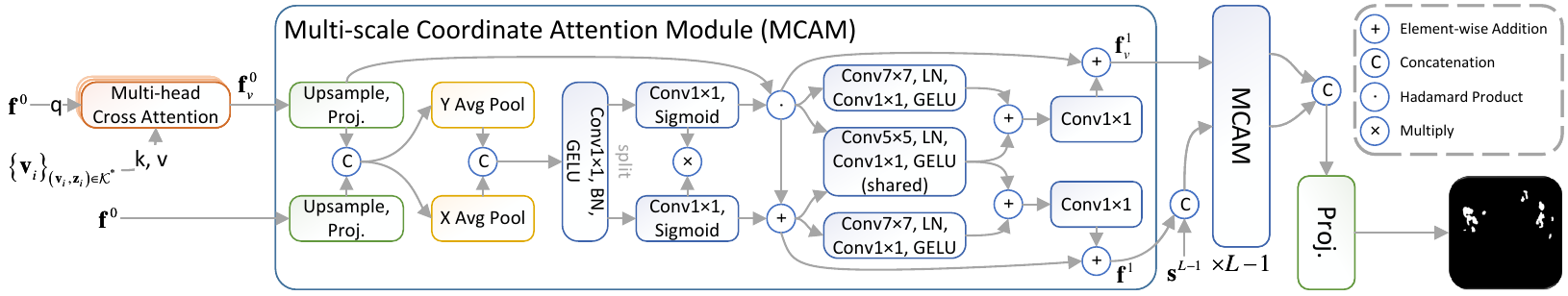}
    \caption{Detailed structure of Multi-scale Coordinate Attention Module.}
    \label{fig:decoder}
\end{figure}
Using the features $\mathbf{f}^0=\mathcal{E}(x,t)$ from the encoder as queries and the retrieved features $\{\mathbf{v}_i\}_{(\mathbf{v}_i,\mathbf{z}_i)\in\mathcal{K}^{\ast}}$ as keys and values enables the model to selectively retrieve and integrate relevant external knowledge guided by the current visual context, thereby enhancing feature discriminability and contextual consistency. It is formulated via multi-head cross attention~\cite{vaswani2017attention} as:
\begin{equation}
\mathbf{f}_{\mathbf{v}}^0=\mathrm{MHCA}(\mathbf{f}^0, \{\mathbf{v}_i\}, \{\mathbf{v}_i\}), \; (\mathbf{v}_i,\mathbf{z}_i)\in\mathcal{K}^{\ast}.
\end{equation}

As depicted in Fig.~\ref{fig:decoder}, Multi-scale Coordinate Attention Module (MCAM) is employed in the decoder to fuse the feature $\mathbf{f}^0$ with $\mathbf{f}_{\mathbf{v}}^0$ in a spatially aware manner. The resulting representation $\mathbf{f}^1$ combined with skip features $\mathbf{s}^{l-1}$ of $l-1^{\mathbf{th}}$ encoder layer and $\mathbf{f}_{\mathbf{v}}^1$ is subsequently fed into next MCAM in $\mathcal{D}_R$. The decoding and fusion for each stage can be expressed as follows:
\begin{equation}
(\mathbf{f}^{i+1}, \mathbf{f}_{\mathbf{v}}^{i+1})=
\mathrm{MCAM}(\mathrm{Concat}(\mathbf{f}^{i}, \mathbf{s}^{L-i}), \mathbf{f}_{\mathbf{v}}^i), \; i\in [1, L-1],
\end{equation}
where $L$ is the number of main stages in the encoder. Inspired by coordinate attention~\cite{hou2021coordinate}, MCAM enhances spatial sensitivity by modeling feature responses along horizontal and vertical directions, which is beneficial for preserving boundary structures during decoding. In addition, multi-scale feature processing allows the module to capture both local details and global context, improving the effectiveness of reference feature integration.


\subsection{Knowledge-Guided Learning Objective}
\subsubsection{Knowledge Update.}

The knowledge pool $\mathcal{K}$ stores a visual prototype $\mathbf{v}_i^{\mathcal{K}}$ and a CoT prototype $\mathbf{z}_i^{\mathcal{K}}$ for each labeled sample $i$. The visual prototype are updated during training via exponential moving average (EMA). Concretely, given the current pooled image feature $\bar{\mathbf{f}}_i=\mathcal{E}(x_i^l)$, the pool entries are updated as: 
\begin{equation}
    \mathbf{v}_i^{\mathcal{K}}\leftarrow \alpha\,\mathbf{v}_i^{\mathcal{K}} + (1-\alpha)\,\bar{\mathbf{f}}_i,
\end{equation}
 where $\alpha\in[0,1)$ is the EMA momentum. Updates are performed only for labeled entries, and stored prototypes are detached from the gradient graph to stabilize retrieval and allow inexpensive persistence.

\subsubsection{Total Loss Function.}
For unlabeled samples we enforce cross-branch consistency between the two decoder outputs. The consistency loss is written as:
\begin{equation}
    \mathcal{L}_{\mathrm{cons}}=\frac{1}{|\mathcal{S}_u|}\sum_{x\in\mathcal{S}_u}\big\|\mathcal{D}_S\left(\mathcal{E}(x, t)\right)-\mathcal{D}_R\left(\mathcal{E}(x, t),\{\mathbf{v}_i\}_{(\mathbf{v}_i,\mathbf{z}_i)\in\mathcal{K}^{\ast}}\right)\big\|_2^2.
\end{equation}

Overall training objective is by weighting supervised and consistency terms:
\begin{equation}
\mathcal{L}_{\mathrm{total}}=\mathcal{L}_{\mathrm{sup}}+\lambda_{\mathrm{cons}}\mathcal{L}_{\mathrm{cons}}.
\end{equation}
 Notably, retrieval affects only the output of $\mathcal{D}_R$, so the consistency term encourages the base decoder $\mathcal{D}_S$ to absorb useful retrieval-induced cues without directly exposing $\mathcal{D}_S$ to retrieved features.

\section{Experiments}
\subsection{Datasets and Experimental Details}
\subsubsection{Datasets.}
We conducted experiments on three public datasets, covering different modalities.
\begin{enumerate*}[label=(\arabic*)]
    \item MosMedData+~\cite{morozov2020mosmeddata} consists of 2\,729 chest CT scans with COVID-19 infection annotations.
    \item QaTa-COV19~\cite{degerli2022osegnet} contains 9\,258 chest X-ray images with COVID-19 infection segmentation masks.
    \item BRISC 2025~\cite{fateh2025brisc} consists of 4\,793 brain MRIs with annotated tumor regions and lesion masks.
\end{enumerate*}
   We split each dataset into train/val/test with an 8:1:1 ratio. The textual annotations for MosMedData+ and QaTa-COV19 follow LViT~\cite{li2023lvit}, while BRISC 2025 annotations were auto-generated from metadata and graphical features.

\subsubsection{Implementation Details.}
All experiments including both training and evaluation phases were conducted on a single NVIDIA RTX 4090 GPU. We set $\lambda_{\text{clip}}=0.5$ and $\alpha=0.99$. The $\lambda_{\text{cons}}$ was initialized as 0.003 and increased via a standard sigmoid ramp-up. For fair comparison, all methods adopted ConvNeXt~\cite{liu2022convnet} as the backbone and CXR-BERT~\cite{boecking2022making} as CoT encoder whenever applicable. Each model was re-implemented and trained for 200 epochs under the same training protocol. The batch size was set to 16, and optimization was performed using SGD with an initial learning rate of $1e^{-2}$ and a weight decay of $1e^{-4}$. Performance was evaluated using Dice and IoU.

\subsection{Comparison with State-of-the-Art Methods}
\begin{table}[htbp]
  \centering
  \caption{Quantitative comparison in terms of Dice (\%) and IoU (\%) on MosMedData+, QaTa-COV19 and BRISC 2025 with 50\% and 25\% labeled ratio. \textbf{Bold} and \underline{underlined} entries indicate the best and second-best results respectively.}
  \label{tab:comparison_25_50}

    \begin{tabular}{lclccccccc}
    \toprule
    \multirow{2}{*}{Method} & \multirow{2}{*}{Text} & \multirow{2}{*}{Backbone} & \multirow{2}{*}{Labeled} & \multicolumn{2}{c}{MosMedData+} & \multicolumn{2}{c}{QaTa-COV19} & \multicolumn{2}{c}{BRISC 2025}\\
    \cmidrule(lr){5-6} \cmidrule(lr){7-8} \cmidrule(lr){9-10}
          &  &  &  & Dice  & IoU  & Dice  & IoU & Dice & IoU \\
    \midrule
    U-Net~\cite{ronneberger2015u} &  & ConvNeXt & 100\% & 76.99 & 66.03 & 75.70 & 64.90 & 82.57 & 74.21\\
    Ours  & $\checkmark$ & ConvNeXt & 100\% & 79.21 & 68.53 & 81.40 & 71.60 & 84.40 & 76.28 \\
    \midrule
    BCP~\cite{bai2023bidirectional}   &  & ConvNeXt & 50\% & 72.57 & 60.79 & 72.59 & 60.92 & 76.66 & 67.42 \\
    DuCiSC~\cite{wu2025dual} &  & ConvNeXt & 50\% & 74.24 & 63.08 & 71.51 & 59.84 & 78.85 & 70.67 \\
    LeFeD~\cite{zeng2024consistency} &  & V-Net~\cite{milletari2016v} & 50\% & 74.77 & 63.38 & 74.64 & 63.74 & 79.63 & 72.13 \\
    LViT~\cite{li2023lvit}  & $\checkmark$ & LViT  & 50\%  & 75.47 & 63.27 & \underline{79.08} & \underline{69.65} & \underline{83.85} & \underline{75.88} \\
    MT~\cite{tarvainen2017mean}    &  & ConvNeXt & 50\%  & 73.44 & 61.97 & 73.39 & 62.22 & 76.38 & 67.88 \\
    UCMT~\cite{shen2023co}  &  & ConvNeXt & 50\%  & \underline{75.95} & \underline{64.65} & 74.42 & 63.33 & 80.04 & 71.70\\
    Ours  & $\checkmark$ & ConvNeXt & 50\%  & \textbf{78.26} & \textbf{66.87} & \textbf{81.23} & \textbf{71.42} & \textbf{84.03} & \textbf{76.03} \\
    \midrule
    BCP~\cite{bai2023bidirectional}   &  & ConvNeXt & 25\%  & 71.31 & 59.37 & 72.44 & 60.81 & 72.46 & 62.43 \\
    DuCiSC~\cite{wu2025dual} &  & ConvNeXt & 25\%  & 71.67 & 60.30 & 70.20 & 58.22 & 76.38 & 68.04 \\
    LeFeD~\cite{zeng2024consistency} &  & V-Net~\cite{milletari2016v} & 25\%  & 70.45 & 58.95 & 74.31 & 63.61 & 78.64 & 71.01 \\
    LViT~\cite{li2023lvit}  & $\checkmark$ & LViT & 25\% & 71.16 & 58.13 & \underline{77.48} & \underline{67.30} & \textbf{80.51} & \textbf{72.98} \\
    MT~\cite{tarvainen2017mean}    &  & ConvNeXt & 25\%  & 70.08 & 58.34 & 72.86 & 61.48 & 71.12 & 62.42 \\
    UCMT~\cite{shen2023co}  &  & ConvNeXt & 25\%  & \underline{72.00} & \underline{60.47} & 72.15 & 60.69 & 74.32 & 66.19 \\
    Ours  & $\checkmark$ & ConvNeXt & 25\%  &\textbf{74.73}&\textbf{63.39}& \textbf{80.52} & \textbf{70.52} & \underline{80.34} & \underline{71.80} \\
    \bottomrule
    \end{tabular}%
\end{table}
\begin{figure}[htbp]
  \centering
    \includegraphics[width=\linewidth]{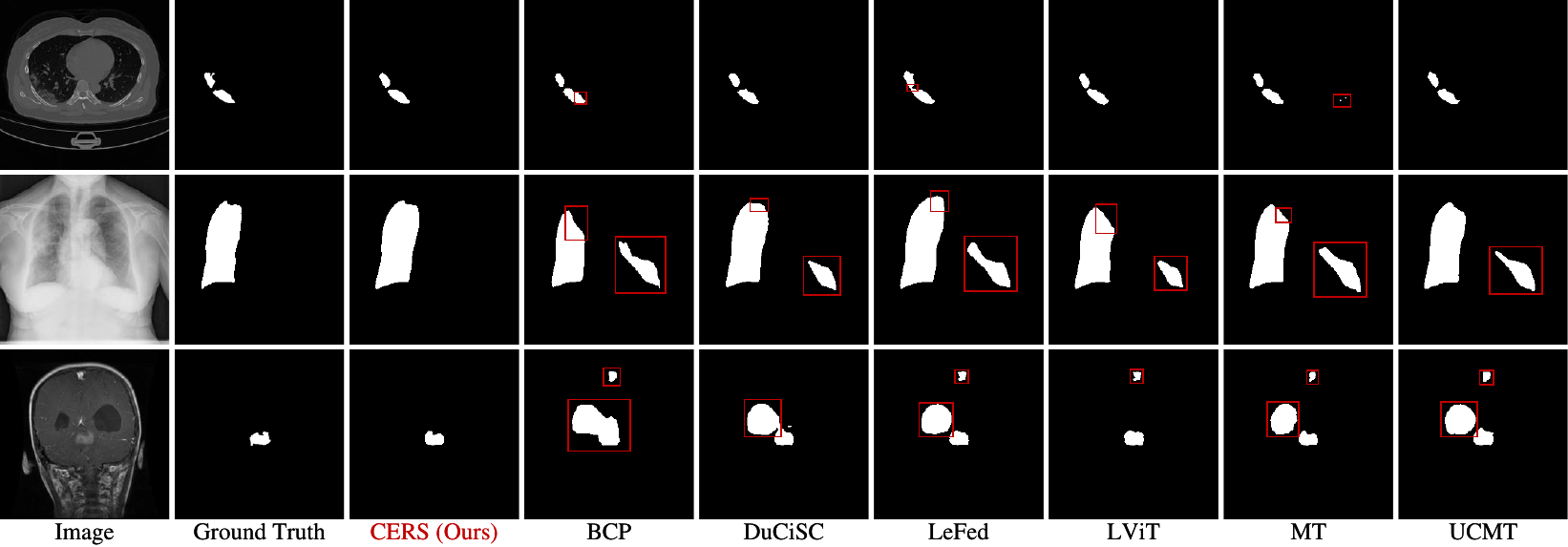}
    \caption{Exemplar qualitative results of different approaches on  MosMedData+ (row 1), QaTa-COV19 (row 2) and BRISC 2025 (row 3).}
    \label{fig:example}
\end{figure}
We compare against six SOTA SSL baselines: BCP~\cite{bai2023bidirectional}, DuCiSC~\cite{wu2025dual}, LeFeD~\cite{zeng2024consistency}, LViT~\cite{li2023lvit}, MT~\cite{tarvainen2017mean} and UCMT~\cite{shen2023co}. We evaluate semi-supervised performance under 50\% and 25\% labeled ratios across three public datasets. The quantitative comparison results are presented in Table~\ref{tab:comparison_25_50}. Compared with existing methods, CERS consistently outperforms SOTA methods in most metrics and datasets. 

Fig.~\ref{fig:example} provides qualitative examples where CERS better localizes lesions and preserves boundaries under ambiguous, low-contrast, artifact-corrupted, or visually confusing cases—demonstrating the value of moving beyond purely visual cues. Unlike pseudo-labeling or teacher-updating methods that suffer from confirmation bias, CERS mitigates noisy supervision by leveraging image–CoT joint cues. Our method remains effective even under fully supervised learning, significantly outperforming the U-Net~\cite{ronneberger2015u} model that employ the same backbone.

\subsection{Ablation Study}
\subsubsection{Impact of Different Retrieval.}

Table~\ref{tab:retrieval} presents the results of different search strategies on the MosMedData+ and QaTa-COV19 datasets with a 50\% labeled ratio. The retrieval strategy provides a reference for the model to perform segmentation. Image retrieval defines a coarse boundary for the search space, while CoT further bridges the gap between visual appearance and diagnostic logic.
\begin{table}[tbp]
  \centering
  \caption{Ablation study on different retrieval strategies of CERS. Latency refers to the time overhead incurred per training step due to the retrieval computation.}

  \label{tab:retrieval}
    \begin{tabular}{cccccccr}
    \toprule
    \multirow{2}{*}{\shortstack{Text\\Retrieval}} & \multirow{2}{*}{\shortstack{CoT\\Retrieval}} & \multirow{2}{*}{\shortstack{Image\\Retrieval}} 
    &\multicolumn{2}{c}{MosMedData+} & \multicolumn{2}{c}{QaTa-COV19} & \multirow{2}{*}{\shortstack{Latency\\per Step}}\\
    \cmidrule(lr){4-5} \cmidrule(lr){6-7}
        &    &    & Dice  & IoU   & Dice  & IoU & \\
    \midrule
        &   &   & 73.51 & 62.36 & 74.07 & 62.86 & 0.0\\
        $\checkmark$ &   &  & 76.68 & 65.54 & 79.43 & 69.13 & +0.048s\\
        & $\checkmark$ &   & \underline{77.22} & \underline{66.07} & \underline{80.86} & \underline{71.05} & +0.055s \\
        &   & $\checkmark$ & 73.60 & 62.14 & 80.66 & 70.66 & +0.047s \\
        $\checkmark$&   &   $\checkmark$ & 77.07 & 65.90 & 80.20 & 70.16 & +0.099s\\
        & $\checkmark$ &   $\checkmark$ & \textbf{78.26} & \textbf{66.31} & \textbf{81.23} & \textbf{71.42} & +0.103s\\
    \bottomrule
    \end{tabular}%
\end{table}
\subsubsection{Impact of Top-$K$ on Different Labeled Data Ratios.}
The performance of different numbers of CoT retrieval cues under varying labeled ratios is illustrated in Fig.~\ref{fig:topk}. Utilizing too few cues fails to provide a robust reference for the model, whereas an excessive number increases the risk of introducing noise. For the MosMedData+ dataset, $K=5$ proves to be a reasonable choice.

\begin{table}[bp]
  \centering
  \begin{minipage}[tb]{0.55\textwidth}
    \centering
    \includegraphics[width=\linewidth]{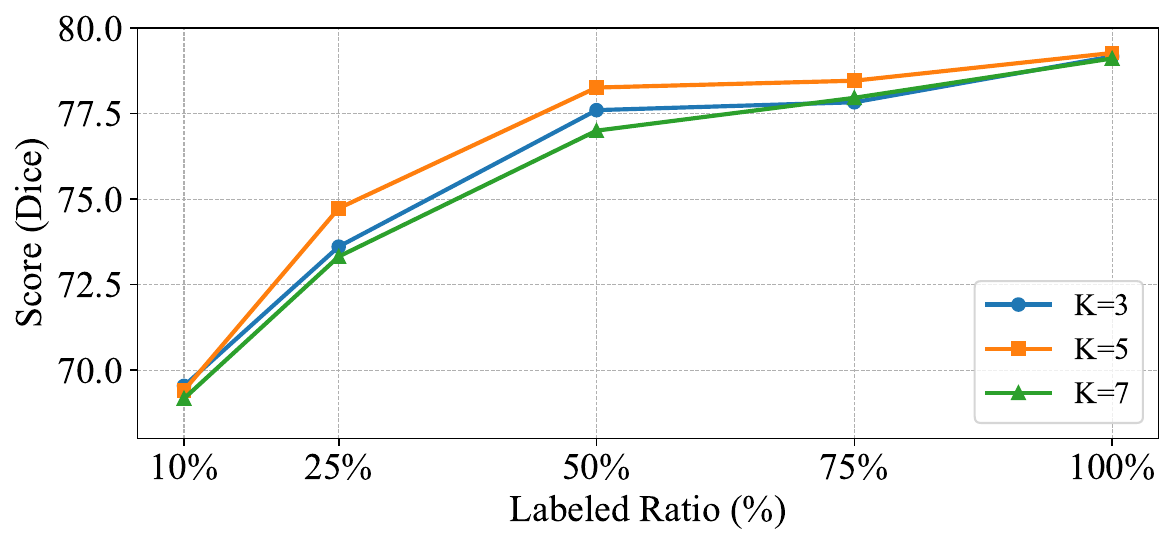}
    \makeatletter\def\@captype{figure}\makeatother
    \caption{CERS performance under different labeled ratios and top-$K$ on MosMedData+.}
    \label{fig:topk}
  \end{minipage}\hfill
  \begin{minipage}[tb]{0.44\textwidth}
    \centering
    \caption{Ablation study on different decoders to fuse cues.}
    \label{tab:decoder}
    \begin{tabular}{lcccccc}
    \toprule
    \multirow{2}{*}{Decoder}  & \multirow{2}{*}{Total Param} & \multicolumn{2}{c}{BRISC 2025} \\
   \cmidrule(lr){3-4}
       &   & Dice  & IoU \\
    \midrule
    U-Net& 51.36M & 82.58 &  74.68 \\
    Swin U-Net& 57.09M & 79.22 & 70.26 \\
    Ours& 48.49M & \textbf{84.03} & \textbf{76.03} \\
    \bottomrule
    \end{tabular}%
  \end{minipage}
\end{table}

\subsubsection{Impact of MCAM.}
We replace the MCAM with a standard U-Net~\cite{ronneberger2015u} decoder and a Swin U-Net~\cite{cao2022swin} decoder, with the results presented in Table~\ref{tab:decoder}. 
As demonstrated in Fig.~\ref{fig:example}, MCAM effectively preserves boundary structures and generates smoother, more precise segmentation masks compared to other decoders. To further validate the internal design of MCAM, we conduct ablation studies by replacing its sub-components with 1×1 convolutions. Specifically, removing the Coordinate Attention module causes a 1.52\% Dice drop, while removing the Multi-scale Branch leads to a 0.95\% drop, confirming the individual effectiveness of both modules within MCAM. Moreover, MCAM efficiently fuses and leverages the image--CoT joint cues without introducing a number of additional parameters to the overall model.

\section{Conclusion}
In this paper, we present CERS, a CoT-enhanced semi-supervised segmentation framework that moves beyond purely visual supervision by incorporating reasoning-driven semantic guidance. Our results suggest that introducing structured diagnostic reasoning helps distinguish visually similar but semantically different regions, improving robustness in challenging scenarios where visual cues alone are insufficient. This finding highlights the importance of high-level semantic reasoning for medical image segmentation, especially under limited annotation settings. Despite its effectiveness, CERS relies on the quality of generated CoTs and an external retrieval mechanism, which may introduce additional computational overhead and dependency on the reasoning model. Future work will explore more efficient reasoning representations and end-to-end integration with medical foundation models to further improve scalability and generalization.

\begin{credits}
\subsubsection{\ackname} This work was supported by the National Natural Science Foundation of China (Grant No 62476054 and Grant No 62576153).

\subsubsection{\discintname}
The authors have no competing interests to declare that are relevant to the content of this article.
\end{credits}

%
%
%
\bibliographystyle{splncs04}
\bibliography{Paper-1650}
%




\end{document}